\documentclass[runningheads]{llncs}

\usepackage[T1]{fontenc}
\usepackage{graphicx}
\usepackage{amsmath,amssymb}
\usepackage{booktabs}
\usepackage{float}
\usepackage{url}
\usepackage{multirow}
\usepackage{adjustbox}
\usepackage[table]{xcolor}
\usepackage{caption}
\begin{document}

\title{PBLH Estimation from Satellite Radiances via a Dual-Encoder Transformer}

\titlerunning{PBLH Estimation via Dual-Encoder Transformer}

\author{Lorenzo Innocenti\inst{1,2} \and
        Luca Catalano\inst{2} \and
        Edoardo Arnaudo\inst{2} \and
        Claudio Rossi\inst{2} \and
        Salvatore Larosa\inst{3,4} \and
        Domenico Cimini\inst{3} \and
        Paolo Garza\inst{1}
        }

\authorrunning{L. Innocenti et al.}

\institute{
    Politecnico di Torino, Torino, Italy\\
    \email{lorenzo.innocenti@polito.it}, \email{paolo.garza@polito.it} \and
    Fondazione LINKS, Torino, Italy\\
    \email{luca.catalano@linksfoundation.com},
    \email{edoardo.arnaudo@linksfoundation.com},
    \email{claudio.rossi@linksfoundation.com} \and
    Institute of Integrated Methodologies for Earth Observation, National Research Council (IMIOT/CNR), Potenza, Italy\\
    \email{salvatore-larosa@cnr.it}, \email{domenico.cimini@cnr.it} \and
    University of Calabria, Arcavacata di Rende, Italy
}

\maketitle

\begin{abstract}
\begin{sloppypar}
Estimating the Planetary Boundary Layer Height (PBLH) from satellite observations is a challenging regression problem due to the indirect relationship between top-of-atmosphere radiances and near-surface atmospheric structure. Progress has been limited both by the lack of architectures capable of handling the multimodal, spatially incomplete nature of satellite overpasses, and by the scarcity of suitable datasets. In this paper, we build upon the large-scale dataset pairing MetOp radiances with ERA5 PBLH labels that we introduced in our previous work, making three contributions. First, we establish a benchmark across eight approaches spanning pixel-wise regression, swath-wise sequence models, and convolutional and Transformer models operating on the full orbital passage. Second, we quantify what the resulting model actually relies on, using grouped Shapley decomposition over the input blocks. Third, we present the best-performing architecture found: a dual-encoder Transformer whose masked-input handling lets it operate in all weather conditions. The proposed model achieves MAE\,=\,155.8~m on the held-out global test set, outperforming all baselines on every evaluation subset. On 30 out-of-distribution granules acquired on two days overlapping the TEAMx observational campaign, it achieves MAE\,=\,165.3~m, outperforming a pixel-wise baseline trained on the same data (MAE\,=\,197~m).
\end{sloppypar}

\end{abstract}

\section{Introduction}

The Planetary Boundary Layer (PBL) is the turbulent lowest layer of the troposphere, directly coupled to the Earth's surface through the exchange of heat, moisture, and momentum. A fundamental parameter characterising the PBL is its height (PBLH), which sets the vertical volume available for the dilution of surface emissions and the exchange of energy between the surface and the free troposphere, with direct relevance to air quality, wind energy, and weather prediction.

Despite this importance, PBLH monitoring relies predominantly on sparse in-situ networks (radiosonde launches, ceilometers, lidars) or on numerical weather models, which are slow and computationally heavy to run. On the other hand, satellite remote sensing can enable continuous, globally dense PBLH mapping at each orbital overpass, dramatically expanding observational coverage \cite{Cimini2024}. Progress toward this goal has been limited both by the lack of architectures capable of jointly handling the heterogeneous, spatially incomplete nature of multi-instrument overpasses, and by the scarcity of large-scale, publicly available datasets pairing satellite radiances with reliable PBLH labels. Building on the large-scale dataset pairing MetOp satellite radiances with ERA5 PBLH labels introduced in \cite{larosa2026pablh}, we propose three new contributions.
\begin{enumerate}
    \item \textbf{A comparative benchmark} across eight models spanning pixel-wise regression, swath-wise sequence models, and convolutional and Transformer architectures operating on the complete passage, establishing the first comparison of this kind for satellite-based PBLH estimation, reported with parameter counts and per-pixel inference cost so that accuracy can be weighed against computational footprint.
    \item \textbf{An attribution analysis of what the model relies on}, using grouped Shapley decomposition over the 13 input blocks, computed exhaustively rather than sampled and scored directly in metres of RMSE.
    \item \textbf{A dual-encoder Transformer architecture} for PBLH estimation from complete MetOp orbital passages, which attains the best accuracy in the benchmark. The model routes infrared (IASI) and microwave (AMSU-A, MHS) radiances through two parallel Vision Transformer encoders, fuses them via element-wise addition, and passes the result through a Transformer decoder to produce dense per-pixel PBLH maps. A dynamic masking schedule, used here as a training-time augmentation that mimics the gaps real passages contain (cloud contamination of the infrared channels, but also fill-flagged or invalid values) rather than a self-supervised reconstruction objective, allows it to consume passages with such gaps directly, in contrast to the convolutional and sequence baselines, which require the missing values to be imputed first.
\end{enumerate}
The proposed model outperforms the pixel-wise baseline of \cite{larosa2026pablh} on the same data and compares favourably with prior PBLH retrieval methods, which rely predominantly on lidar backscatter or reanalysis predictors rather than on passive sounder radiances \cite{ayazpour2023pblh,salcedo2025calipso,li2026cats}. Code and dataset are available at \url{github.com/links-ads/pblh-transformer}.

\section{Related Work}

The PBL was recommended as an Incubation Targeted Observable by the 2017 US Decadal Survey for Earth Science and Applications from Space (ESAS 2017)~\cite{nasem2018decadal}, a tier reserved for observables judged scientifically important but not yet ready for a dedicated mission. A subsequent NASA-commissioned study~\cite{nasa2023pblincubation} set out the associated measurement requirements, roughly 1--25~km horizontal and 0.2--2~km vertical resolution at 1--24~h revisit for temperature and water vapour profiling, and assessed the observational gaps left by the current program of record. Every satellite-based PBLH estimate in the literature, the present work included, is consequently derived from instruments designed for other purposes: lidar backscatter intended for aerosol and cloud profiling, GNSS radio occultation intended for numerical weather prediction refractivity, or infrared and microwave sounders intended for atmospheric temperature and humidity retrieval.

Operational PBLH profiling relies primarily on radiosonde networks and ground-based remote sensing instruments such as lidars, ceilometers, and microwave radiometers (exemplified by the pan-European E-PROFILE network \cite{Rufenacht2021}), which provide continuous time series at instrumented sites but leave vast ocean, polar, and developing-world regions unobserved. 
Satellite observations overcome the coverage limitation of ground stations by offering a global perspective at each overpass, and PBLH estimation from space has been pursued by applying the same gradient-detection principles to retrieved atmospheric profiles. Zhang et al.~\cite{zhang2016caliop} estimate PBLH from CALIPSO lidar backscatter over China, reporting correlations of $R=0.59$ at Beijing and $R=0.65$ at Jinhua against ground-based lidar. GNSS radio occultation offers a complementary space-based route, resolving the refractivity gradient at the PBLH top with a vertical resolution of a few hundred metres, but its horizontal footprint is too coarse for sharp gradients and its sampling is sparse and irregular~\cite{su2026pbl}.

However, the indirect and non-linear relationship between top-of-atmosphere radiances and near-surface atmospheric structure makes retrieval challenging, motivating a shift toward machine learning approaches that learn the mapping directly from data. Almost all of this effort rests on active sensing or on reanalysis rather than on passive radiances. Salcedo-Bosch et al.~\cite{salcedo2025calipso} apply a Random Forest to ten years of CALIOP backscatter profiles paired with radiosonde measurements, achieving an RMSE of 278~m over Europe and North America, and Li et al.~\cite{li2026cats} apply an attention-augmented ResNet with transfer learning to CATS backscatter profiles augmented with reanalysis fields, achieving an MAE of 561~m and $R=0.67$ against radiosondes. Ayazpour et al.~\cite{ayazpour2023pblh} instead forgo satellite radiances altogether, training XGBoost on ERA5 meteorological and geographical predictors against AMDAR aircraft profiles to produce spatially complete PBLH over the contiguous United States, with a cross-validated MAE of 186~m at held-out airports.

Passive infrared and microwave sounders sit at the opposite end of this trade-off: they sample a wide swath at high revisit and, in the microwave, retain sensitivity through cloud. Retrieval of PBLH from sounder radiances alone remains largely unexplored; the closest work is our own \cite{larosa2026pablh}, which estimates PBLH from combined infrared and microwave radiances footprint by footprint. The only other work to combine the two modalities is Milstein et al.~\cite{milstein2023detail}, who apply a 3D deep neural network to enhance the vertical detail of existing AIRS/AMSU Level-2 retrieved temperature and humidity profiles; their model operates on already-retrieved profiles rather than raw radiances and does not target PBLH directly, so its reported accuracy is not directly comparable to ours. A further limitation common to all of these approaches is that they treat each retrieval location independently, discarding the spatial structure of the orbital passage. To our knowledge, the present work is the first to estimate PBLH directly from raw infrared and microwave radiances in a dual-encoder architecture and to leverage spatial context across the full satellite overpass. On this dataset, exploiting horizontal context lets the model reach an accuracy competitive with lidar- and reanalysis-based approaches despite the coarse vertical resolution of passive sounding, a result we see as an encouraging first step rather than a final answer on how broadly MetOp-class radiances can support PBLH retrieval.

\section{Dataset}

We use the dataset introduced in \cite{larosa2026pablh}, which pairs raw top-of-atmosphere radiances from two complementary satellite sensors with ERA5 PBLH targets. Full details of how the dataset was constructed are given in \cite{larosa2026pablh}. The input radiances come from IASI and AMSU-A/MHS aboard the MetOp satellites (A, B, C), which are freely and openly distributed by EUMETSAT. Each MetOp satellite follows a polar sun-synchronous orbit, providing approximately two overpasses per day over any location; the three-satellite constellation ensures consistent global sampling. The dataset covers the year 2022, sampled at weekly intervals (one date per week from January 3 to December 26), with four synoptic overpass times per date (00, 06, 12, and 18 UTC), yielding 194 orbital passages in total and over 4.3 million samples distributed across all seasons and latitudes.
The target PBLH is derived from the ERA5 fifth-generation reanalysis \cite{hersbach2020era5}. Each satellite footprint's ERA5 PBLH target was obtained by sampling the four nearest grid points and taking their mean.

For the out-of-distribution evaluation we additionally use 30 MetOp granules acquired on two dates outside the 2022 dataset, 14 February 2025 and 25 June 2025 (15 granules each), selected to fall within the observation period of the TEAMx campaign.\par
The dataset is partitioned to prevent data leakage while ensuring all seasons are represented in every split. Within each month, the first sampled date is assigned to the test set, the second to validation, and all remaining dates to training. This is applied independently for each month, so test and validation each contain one day per month spread across the full annual cycle, while training retains the majority of the data.
Each channel is independently standardised using per-channel mean and standard deviation computed from the training set. The PBLH target is normalized using a fixed scale factor of 4000~m.
In addition to the spectral channels, each observation is augmented with auxiliary features. Periodic quantities (latitude, longitude, time of day, day of year, satellite and solar angles) are sine/cosine encoded with period equal to their natural range. Surface elevation, cloud flag, and land flag are also included. The final input vector has $C = 207$ channels per observation: 20 microwave brightness temperatures, 168 infrared radiances, and 19 auxiliary features.

\begin{figure}[t]
    \centering
    \includegraphics[width=\textwidth]{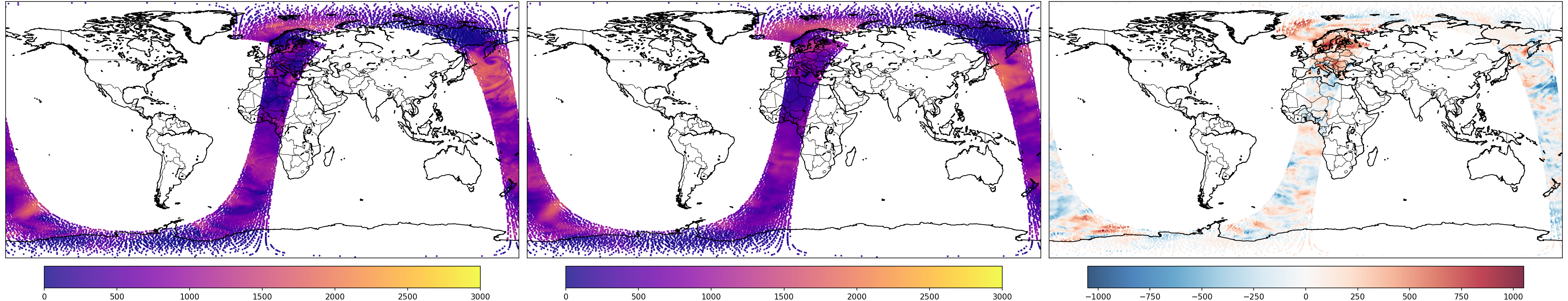}
    \caption{PBLH along the MetOp ground tracks for the overpasses of 14 February 2025. Left: ERA5 PBLH, used as the reference. Centre: Transformer-2D prediction. Right: prediction minus reference. The first two panels share a common 0--3000~m colour scale; the difference panel uses a diverging scale spanning $\pm$1000~m.}
    \label{fig:teamx}
\end{figure}

\section{Methodology}
\label{sec:methodology}

We compare seven baseline models of increasing spatial complexity against our proposed Transformer-2D architecture. Models are grouped into three spatial granularities, each exploiting a different level of context from the satellite data.

\subsection{Pixel-wise Models}

Pixel-wise models treat each IASI footprint independently, mapping a single $\mathbf{x} \in \mathbb{R}^{207}$ feature vector to a scalar PBLH estimate with no spatial context.
A ridge-regularised linear model, Random Forest, and XGBoost serve as non-deep baselines of increasing complexity.
Multilayer Perceptron (MLP) is a fully connected neural network built from residual blocks \cite{he2016deep}, each comprising two linear transformations with batch normalisation and ReLU activations, used as a deep learning baseline for pixel-wise regression.

\subsection{Swath-wise Models}

Swath-wise models process one full scan line at a time, treating the $W = 30$ cross-track positions as a 1D sequence. Missing values are filled by 1D linear interpolation so that the model always receives a complete sequence.

UNet-1D is a 1D encoder-decoder with skip connections \cite{ronneberger2015u}, progressively downsampling and upsampling the swath.
Transformer-1D applies self-attention along the scan line \cite{vaswani2017attention}, with positional encodings to preserve scan order.

\subsection{Passage-wise Models}

\begin{figure}[t]
    \centering
    \includegraphics[width=\textwidth]{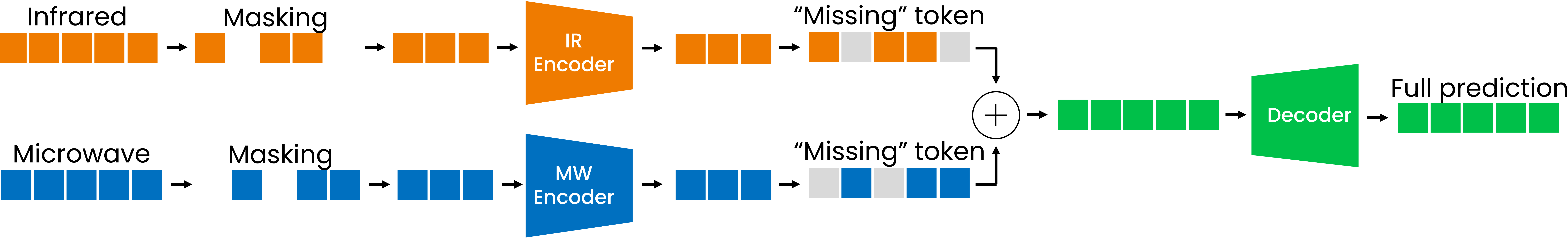}
    \caption{Overview of the proposed dual-encoder Transformer. Infrared and microwave radiances are tokenised and processed by two independent ViT encoders. The encoded representations are fused by element-wise addition; learnable missing-token embeddings are inserted at masked positions before a shared Transformer decoder produces per-pixel PBLH estimates.}
    \label{fig:architecture}
\end{figure}

Passage-wise models operate on complete 2D orbital passages, treating the overpass as a spatial image of shape $H \times W \times C$.

ResNet-2D requires a fully dense input since convolutional kernels cannot process sparse missing values. Missing pixels are therefore filled by 2D bilinear interpolation prior to model input, with any remaining gaps resolved iteratively using the mean of valid 8-connected neighbours. During training, passages are augmented with random horizontal and vertical flips, 90° rotations and random affine transforms (shift, scale, small rotation). A random crop of size $P \times P$ is then applied. At inference, the full passage is fed to the network after symmetric mirror padding so as to avoid border artifacts.

\subsubsection{Proposed approach} 

Transformer-2D is our proposed model. It addresses the limitations of feature-based regression and convolutional networks through a multimodal dual-encoder architecture that handles heterogeneous satellite inputs and missing observations via a masking-based augmentation strategy adapted from the masked-autoencoder architecture \cite{he2022mae}. Random input masking mirrors real acquisition gaps at training time, so the model learns to recover from them directly through the downstream regression loss, rather than via a separate reconstruction objective or imputation.
The 207-channel input is divided into two sensor streams processed by dedicated encoders before fusion. The microwave stream comprises the 20 brightness temperature channels concatenated with the 19 auxiliary features (39 channels total). The infrared stream comprises the 168 radiance channels concatenated with the same auxiliary features (187 channels total). Both encoders share their structural design but are decoupled in weight space, enabling each stream to learn modality-specific representations.

The full satellite passage (padded to $800 \times 30$ pixels) is processed in the same fashion as the Vision Transformer \cite{dosovitskiy2020image} applied to an image, with each pixel acting as an individual token (patch size $1 \times 1$). Tokens are grouped into localised windows of adjacent samples: pixels in close spatial proximity share high correlation, so that when observations are corrupted or missing, neighbouring pixels carry context that guides recovery of the underlying state. Each encoder processes only the visible (non-masked) tokens in its stream and projects each pixel to a shared latent dimension.
After encoding, the two encoder outputs are fused via element-wise addition. Learnable missing-token embeddings are then inserted at all masked positions, reconstructing a complete token sequence. The decoder mirrors the encoder structure and processes the complete set of tokens, both observed and reconstructed, allowing it to propagate context from visible pixels to masked positions. A final linear projection maps each decoder output token to a scalar PBLH estimate.

During training, a dynamic masking schedule randomly selects a masking ratio between 0\% and 100\% of the available pixels in each window. Randomly masked tokens are replaced by a learnable missing-token embedding, borrowing the missing-token mechanism of \cite{he2022mae}. Microwave and infrared missing tokens are learned separately, which encourages the model to exploit cross-modal redundancy and maintain prediction capability when either sensor stream is partly or fully unavailable. During training, windows are augmented with random horizontal flips, vertical flips, and 90° rotations before the random crop to size $P \times P$. At inference, the full passage is decomposed into overlapping $16 \times 16$ tiles using the same window size as training, with a stride of 8 pixels. Each tile is processed independently and the resulting predictions are averaged wherever tiles overlap, yielding a complete prediction map over the passage.
Figure~\ref{fig:architecture} summarises the full dual-encoder architecture.
By design, each orbital passage is processed independently; we return to the implications of this scoping choice in Section~\ref{sec:limitations}.

\section{Experiments}

\subsection{Configurations}

All models share the same 207-channel input. All hyperparameters were found by grid search. The linear baseline is ridge-regularised ($\alpha = 1$). Random Forest used 100 trees, maximum depth 12, and square-root feature subsampling. XGBoost used 100 boosting rounds with MAE objective, tree depth 6, and early stopping patience 10.
Neural networks were trained with the AdamW optimizer using a cosine learning rate schedule from $10^{-5}$ to $10^{-10}$. The MLP has sixteen residual blocks of dimension 128 and batch normalisation trained with batch size 1024. UNet-1D and Transformer-1D use batch size 1024 on swath segments of length 32. UNet-1D has three encoder stages with channel widths 64, 128, and 256, a bottleneck of width 512, and a symmetric decoder; each stage consists of two residual blocks, batch normalisation, ReLU. Transformer-1D has model dimension 256, 8 attention heads, and 4 encoder layers. ResNet-2D uses batch size 16, and training patch $128\times128$. The Transformer-2D uses batch size 16; each ViT encoder has dimension 384, depth 2, 16 attention heads with head dimension 24, and the decoder shares these settings but with depth 4.

\subsection{Quantitative Results}

Table~\ref{tab:results} reports MAE, RMSE, and Pearson $r$ across four evaluation subsets: the full held-out test set, Europe-only, clear-sky, and cloudy-sky. Models are grouped by spatial granularity: pixel-wise, swath-wise, and passage-wise. Transformer-2D outperforms every baseline on all three metrics and all four subsets, indicating that spatial context across the overpass carries information not present at the single-footprint or swath level. Access to that context is not sufficient on its own, however: ResNet-2D, the other passage-wise model, is beaten by the pixel-wise MLP on RMSE over the full test set and by UNet-1D over Europe on both error metrics. Among pixel-wise models, tree ensembles improve on the regularised linear baseline, and the MLP achieves the lowest MAE of the group. The MLP result deserves comment: on the full test set it outperforms both swath-wise models on all three metrics (177.6~m MAE) and approaches ResNet-2D, despite seeing a single footprint at a time with no spatial context whatsoever and using by far the fewest parameters of the neural baselines (0.56~M). A large fraction of the achievable skill is therefore available from the spectral signature of an individual footprint alone, and spatial context, while clearly beneficial, provides a comparatively modest additional margin. All models show higher errors on the Europe-only subset, consistent with the higher ERA5 PBLH variability over that region (std 438~m vs 410~m globally). The Transformer-2D shows a wider advantage in cloudy conditions compared to the other test sets. This pattern is consistent with the dual-encoder design providing a route toward all-weather operation: when cloud cover renders infrared channels unavailable, the microwave encoder continues to provide a complete representation, and the fusion of both streams may help the model compensate for the missing modality. The masking mechanism further contributes by training the model to recover from arbitrary patterns of missing tokens, so inference-time gaps are handled the same way as training-time masks. We note that the swath-wise models and ResNet-2D all require a fully dense input and therefore receive imputed values wherever observations are missing (1D linear interpolation along the scan line for UNet-1D and Transformer-1D, 2D bilinear interpolation for ResNet-2D), whereas Transformer-2D consumes such gaps directly. 

\begin{table}[t]
\caption{MAE (m), RMSE (m), and Pearson $r$ across evaluation subsets. Models grouped by spatial granularity. Bold: best per row. Parameter counts are given only for models whose capacity is a fixed number of weights. The tree ensembles are instead sized by their fitted structure (RF: 100 trees, 566k nodes; XGBoost: 100 trees, 12.6k nodes); the ridge baseline has 207 weights and one bias.}
\label{tab:results}
\centering
{\scriptsize\setlength{\tabcolsep}{4pt}
\begin{tabular}{ll cccc cc cc}
\toprule
& & \multicolumn{4}{c}{\textit{Pixel-wise}} & \multicolumn{2}{c}{\textit{Swath-wise}} & \multicolumn{2}{c}{\textit{Passage-wise}} \\
\cmidrule(lr){3-6}\cmidrule(lr){7-8}\cmidrule(lr){9-10}
Metric & Subset & Ridge & RF & XGB & MLP & UNet-1D & Tr.-1D & ResNet & \textbf{Tr.-2D} \\
\midrule
\multirow{4}{*}{MAE (m)$\downarrow$}
 & Full   & 238.6 & 215.5 & 194.5 & 177.6 & 191.9 & 187.5 & 166.8 & \textbf{155.8} \\
 & EU     & 298.8 & 289.2 & 262.6 & 247.1 & 222.0 & 245.4 & 238.2 & \textbf{212.6} \\
 & Clear  & 220.8 & 198.8 & 175.5 & 157.5 & 204.9 & 191.0 & 149.1 & \textbf{143.2} \\
 & Cloudy & 243.6 & 220.3 & 199.9 & 183.2 & 193.3 & 187.4 & 171.8 & \textbf{159.3} \\
\midrule
\multirow{4}{*}{RMSE (m)$\downarrow$}
 & Full   & 308.4 & 284.1 & 264.6 & 228.7 & 264.1 & 260.5 & 232.3 & \textbf{215.3} \\
 & EU     & 377.4 & 367.7 & 343.8 & 304.7 & 292.2 & 320.0 & 300.4 & \textbf{268.9} \\
 & Clear  & 286.7 & 260.0 & 240.9 & 200.4 & 283.8 & 266.8 & 207.6 & \textbf{198.5} \\
 & Cloudy & 314.4 & 290.6 & 271.0 & 235.3 & 265.8 & 260.3 & 238.4 & \textbf{219.5} \\
\midrule
\multirow{4}{*}{Pearson $r$$\uparrow$}
 & Full   & .662 & .720 & .769 & .805 & .791 & .796 & .826 & \textbf{.850} \\
 & EU     & .501 & .555 & .644 & .707 & .744 & .707 & .750 & \textbf{.774} \\
 & Clear  & .652 & .712 & .758 & .808 & .744 & .775 & .825 & \textbf{.839} \\
 & Cloudy & .665 & .724 & .772 & .804 & .792 & .796 & .826 & \textbf{.852} \\
\midrule
\multicolumn{2}{l}{Parameters}   & 208 & n/a & n/a & 0.56\,M & 4.23\,M & 3.21\,M & 26.9\,M & 14.6\,M \\
\bottomrule
\end{tabular}}
\end{table}

The Transformer-2D was further evaluated on 30 granules acquired on two dates, 14 February 2025 and 25 June 2025 (15 granules each), with all reported metrics computed over the full set of 30. These dates fall outside the training and validation splits and represent contrasting seasonal conditions (winter and early summer), and were specifically selected within the observation period of the TEAMx observational campaign \cite{Lehner2026}.
Evaluated against ERA5 PBLH on these campaign-period overpasses, the model achieves MAE 165.3~m and RMSE 226.6~m, a bit higher than on the global test set (MAE 155.8~m), with essentially unchanged correlation ($r = 0.852$ against $0.850$). For reference, the per-pixel MLP baseline reported in \cite{larosa2026pablh} obtains MAE~$=$~197~m and RMSE~$=$~254~m on the same granules, suggesting that the spatial context captured by the dual-encoder Transformer improves generalization to this out-of-distribution setting. On a single consumer-grade NVIDIA RTX 2080 Ti, Transformer-2D produces a full-passage PBLH map ($765\times30$ footprints) in approximately 0.46~s (50k pixels/s), so a full day of global MetOp coverage can be processed in a couple of minutes on one such card. Figure~\ref{fig:teamx} maps the ERA5 reference, the Transformer-2D prediction and their difference along the ground tracks for one of the two dates.

\subsection{Ablation Study}

Table~\ref{tab:ablation} isolates the contribution of four architectural choices, measured in RMSE on the global all-sky test set. \textit{Single encoder} replaces the dual-encoder design with a single encoder processing all 207 channels jointly, removing the modality-specific inductive bias. \textit{No masking} fixes the additional masking at 0\%. \textit{Masking cap at 50\%} instead still draws a random amount each step, but from 0\% to 50\% rather than the full 0\% to 100\% range used in the full model, reducing the diversity of missing-data scenarios seen during training. \textit{Concat fusion} replaces element-wise addition with concatenation followed by a linear projection. Every variant increases RMSE relative to the full model. Single encoder is the most costly at +4.1~m; the other three variants are comparatively minor and closely clustered, from +1.1~m (concat fusion) to +2.5~m (masking cap at 50\%).

\subsection{Channel Attribution}
\label{sec:attribution}

To quantify how much each input group contributes to accuracy we employ a grouped Shapley value~\cite{shapley1953value} computed directly in RMSE. The 207 channels are partitioned into 13 blocks: the 20 microwave channels, the 168 infrared channels, and eleven single- or dual-channel auxiliary blocks (cloud flag, land flag, surface elevation, day of year, time of day, latitude, longitude, satellite azimuth, satellite zenith, solar azimuth, solar zenith). The 1024 evaluation crops are $16\times16$ windows, matching the model's native input size, drawn at random locations from the test-set passages and kept only if entirely free of missing values, so every combination of blocks is scored on the same fixed set of fully observed pixels. For each such combination, the blocks left out keep their values from an independently drawn reference crop, sampled the same way, instead of their real ones, and the RMSE is computed over the 1024 evaluation crops.

\begin{figure}[t]
\begin{minipage}[t]{0.345\textwidth}
    \vspace{0pt}
    \centering
    \captionof{table}{Ablation study. Global all-sky test-set RMSE (m) with one architectural choice changed at a time, all else held at the full model.}
    \label{tab:ablation}
    \smallskip
    {\scriptsize
    \begin{tabular}{lc}
    \toprule
    Configuration & RMSE $\downarrow$ \\
    \midrule
    Single encoder & 219.4 \\
    No masking & 217.0 \\
    Masking cap at 50\% & 217.8 \\
    Concat fusion & 216.4 \\
    \textbf{Full model} & \textbf{215.3} \\
    \bottomrule
    \end{tabular}}
\end{minipage}
\hfill
\begin{minipage}[t]{0.63\textwidth}
    \vspace{0pt}
    \centering
    \includegraphics[width=0.82\linewidth]{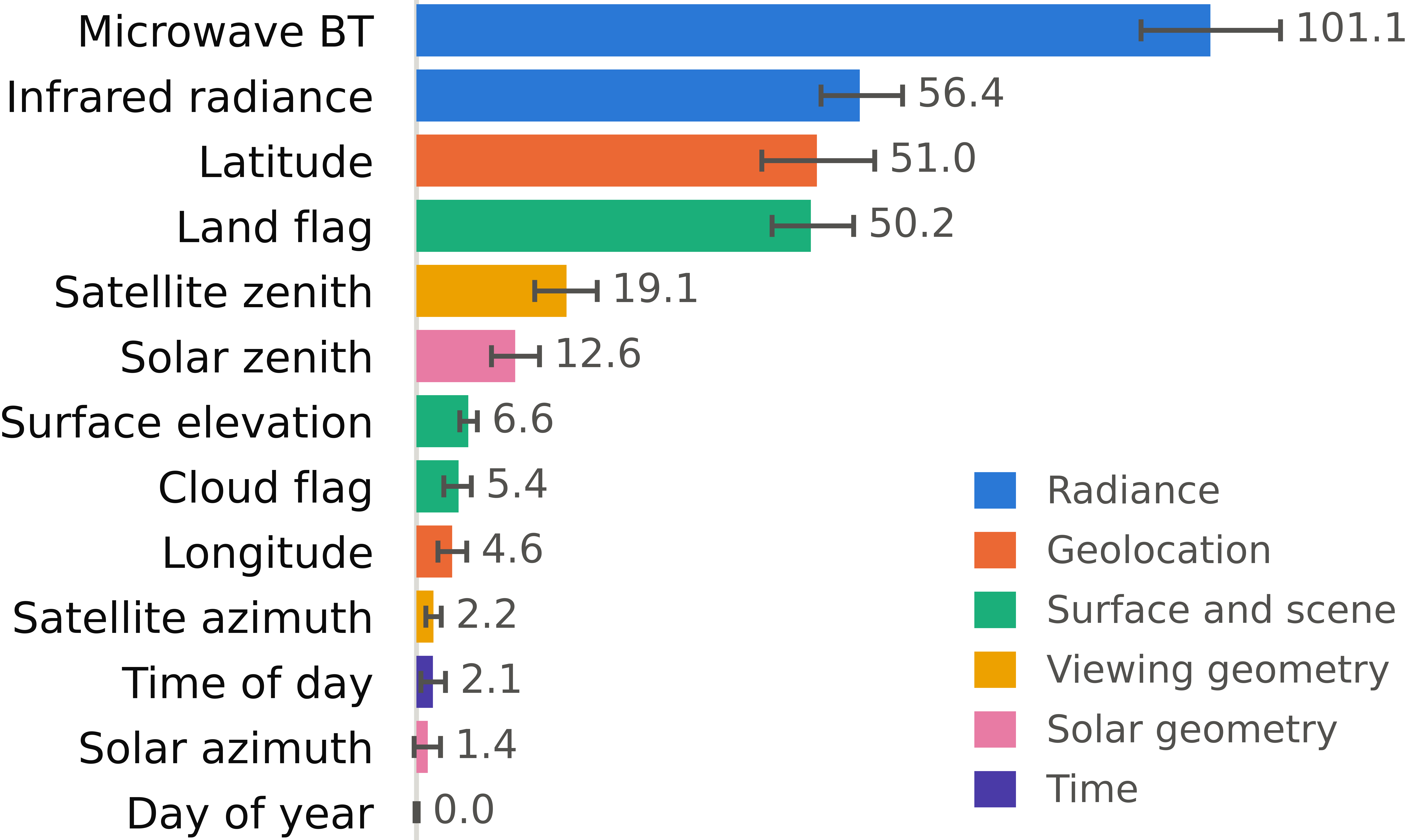}
    \captionof{figure}{Grouped Shapley value per input block, in metres of test-set RMSE reduction. Bars show 95\% bootstrap intervals.}
    \label{fig:shap}
\end{minipage}
\end{figure}

Figure~\ref{fig:shap} reports the Shapley value for every block with 95\% bootstrap confidence intervals. Microwave brightness temperatures carry the single largest share (101.1~m, 32\% of the total), ahead of infrared radiances (56.4~m, 18\%). Latitude (51.0~m) and the land/sea flag (50.2~m) follow closely and are individually comparable in magnitude to the entire 168-channel infrared block, indicating that geolocation and surface-type context carry nearly as much information as the spectral radiances themselves. Satellite zenith and solar zenith contribute more modestly (19.1~m and 12.6~m respectively), and day of year is statistically indistinguishable from zero.

\section{Conclusion}
\label{sec:conclusion}

We presented an exploratory comparative study investigating whether Planetary Boundary Layer Height can be estimated from MetOp satellite radiances alone, across eight machine learning approaches spanning three spatial granularities. Among these, a dual-encoder Transformer achieved the best performance on all evaluation subsets, reaching an MAE of 155.8~m and Pearson $r = 0.850$ on the global held-out test set, including on its clear-sky and cloudy subsets, and degrading only moderately on 30 out-of-distribution granules acquired during the TEAMx observational campaign period.

\subsection{Limitations}
\label{sec:limitations}

Both our training labels and our TEAMx-period evaluation use ERA5 PBLH, a model-derived quantity rather than a direct atmospheric observation. Every number we report therefore measures agreement
with a reanalysis, not with ground measurements, and it is essential to know how good that reanalysis is. Guo et
al.~\cite{Guo2021} evaluate ERA5 PBLH against near-global high-resolution radiosondes and find that ERA5
underestimates daytime PBLH by around 130~m. Our model reaches
MAE~=~155.8~m and $r = 0.850$ against ERA5. In other words, the disagreement between our model and its label is
of the same order as the disagreement between that label and radiosonde observations. Two consequences follow.
First, our error figures cannot be read as distance from the true PBLH; a model in perfect agreement with ERA5
would still carry ERA5's own bias. Second, part of the residual error we report may reflect the label's noise
floor rather than model deficiency, which also means that small differences between the better-performing
models in Table~\ref{tab:results} should be interpreted with caution.

Each orbital passage is processed independently. The model captures spatial context within an overpass but does
not exploit temporal continuity across successive passages, which could help resolve rapidly evolving boundary
layer conditions. This is a deliberate scoping choice rather than an architectural constraint.

\subsection{Future Work}

The most consequential direction follows from the label limitation discussed in Section~\ref{sec:limitations}: replace ERA5 PBLH as the supervision signal with direct ground-based measurements, for instance from in-situ networks such as the pan-European E-PROFILE microwave radiometer and ceilometer network~\cite{Rufenacht2021}, or the TEAMx Observational Campaign once released. This would let the retrieval be evaluated directly against observations and serve as an independent check on the reanalysis itself.

Three further directions are identified for future work. First, extending the dataset to multi-year records would improve the representation of seasonal variability and rare meteorological regimes. Second, incorporating temporally adjacent overpasses as additional context could help the model exploit the temporal continuity of boundary layer evolution and reduce errors in rapidly changing conditions. Third, the model currently outputs a single point estimate per footprint with no associated confidence; adding uncertainty quantification, for instance via quantile regression, ensembling, or conformal prediction, would let predictions be flagged by reliability, which is particularly relevant where inputs are heavily masked or reconstructed and error is expected to be higher.

\begin{credits}
\subsubsection{\ackname}
This work was carried out within the Space It Up! project funded by the Italian Space Agency (ASI) and the Ministry of University and Research (MUR), under contract n.~2024-5-E.0 -- CUP n.~I53D24000060005. This work was also supported by the Twinverse project funded by the European Union under the Horizon Europe programme, grant agreement n.~101270783.

\subsubsection{\discintname}
The authors have no competing interests to declare that are relevant to the content of this article.
\end{credits}

\bibliographystyle{splncs04}
\bibliography{refs}

\end{document}